\title{Lehmer Transform and Its Theoretical Properties}

\author{Masoud Ataei, Shengyuan Chen and Xiaogang Wang \\
	Department of Mathematics and Statistics\\
	York University\\
	Toronto, Ontario , Canada}
\date{}

\documentclass[11pt]{article}

\usepackage{siunitx}
\usepackage{printlen}
\usepackage{epsfig, epstopdf, amsmath, amssymb, graphicx,  booktabs, algorithm, algpseudocode, tikz, lscape, booktabs, mathrsfs, accents, mathtools, subfig, pgf, hyperref, enumerate}
\usepackage[numbers,square,sort]{natbib}
\DeclareMathAlphabet\mathbfcal{OMS}{cmsy}{b}{n}

\usepackage[title]{appendix}

\newcommand{\BlackBox}{\rule{1.5ex}{1.5ex}}  

\newtheorem{theorem}{Theorem}
\newtheorem{lemma}[theorem]{Lemma}

\newtheorem{corollary}[theorem]{Corollary}
\newtheorem{definition}{Definition}

\newcommand{\LT}[3]{\breve{\mathscr{L}}_{#1}\{ \accentset{\circ}{h}_{#2} \}({#3})}
\newcommand{\LTshort}{\breve{\mathscr{L}}}

\newcommand{\LTinverse}[3]{\breve{\mathscr{L}}^{-1}_{#1}\{ \accentset{\circ}{h}_{#2} \}({#3})}

\begin{document}
\maketitle

\begin{abstract}
We propose a new class of transforms that we call {\it Lehmer Transform} which is motivated by the {\it Lehmer mean function}. The  proposed {\it Lehmer transform} decomposes a function of a sample into their constituting statistical moments. Theoretical properties of the proposed transform are presented. This transform could be very useful to provide an alternative method in analyzing non-stationary signals such as brain wave EEG. 
\end{abstract}

\section{Introduction}
In this paper, we introduce a new class of transforms referred to as  the\textit{Lehmer Transform} . The proposed transform decomposes a function of the data  into the so-called \textit{Breve Moments}, which in turn provides the means to construct some parametric families of nonparametric statistical models.

The Lehmer transform is motivated by  the \textit{Lehmer mean function}, named after the renowned number theorist Derrick H. Lehmer \cite{havilgamma}. However, the appealing properties of this function has not received much attention from statistics and machine learning communities in the past.As generalization of the \textit{power mean function} and having connections to other important classes of means, the Lehmer mean function has received a lot of attentions in recent years. The function's elementary properties like homogeneity, monotonicity and differentiability have been discussed in \cite{bullen2013handbook,beliakov2016practical}, whereas its more advanced properties like \textit{Schur-convexity}, \textit{Schur harmonic convexity} and \textit{Schur power convexity} have been the focus of more recent studies \cite{vculjak2012schur, fub2016schur, xia2009schur, yang2013schur, chu2015convexity}. Also, the inflection points of the function have been studied in \cite{sluciak2015inflection}, and the results concerning its possible connections to \textit{Gini} and \textit{Toader} means have been provided in \cite{chu2016sharp, chu2012optimal, trif2002sharp, zhao2017optimal, chu2012inequalities, hua2014double}.
Most application of  the Lehmer mean function  rely mostly on its special cases. For instance, Terziyan \cite{terziyan2017social} formulated a distance metric using the harmonic and contra-harmonic means, and evaluated its effectiveness in geographic information systems. In a similar manner, Somasundaram et al. \cite{somasundaram2016lehmer} studied the disconnected graphs and their possible labeling schemes, and Sluciak \cite{sluciak2014consensus} developed state-dependent consensus algorithms by resorting to some special cases of the Lehmer mean function. In addition, Gomes \cite{gomes2017port} successfully constructed a family of high-performance value at risk estimators through the use of this function and proved asymptotic normality of the constructed estimators.

\section{Discrete Lehmer Transform}
\label{sec:LehmerTransform}
Let $X_1, X_2, \dots$ denote a sequence of random variables defined on some given probability measure space $(\Omega, \mathscr{F}, \mathbb{P})$ such that each random variable takes its value from the same measurable space $(\mathcal{X},\mathscr{B}_{\mathcal{X}})$. Also, let $\mathbf{x}=\left( x_1, x_2, \dots, x_n \right)^\intercal $ denote a realization of some discrete-time stochastic process $\left\{ X_n \right\}$ where no restrictions pertain to its underlying sampling procedure. For instance, $X_1, X_2, \dots, X_n$ could be any combination of $n$ continuous- or discrete-type random variables which are not necessarily independent or identically distributed.

Let us further consider a generic framework to normalize the sample, in which every possible normalization of the vector-valued sample $\mathbf{X}_n = \left( X_1, X_2, \dots, X_n \right)^\intercal $ is viewed as a $(\mathscr{B}_{\mathcal{X}^n},\mathscr{B}_{\mathcal{H}})-$measurable function obtained through composition of several, say $m$, maps; i.e. the function denoted by
\begin{equation*}
\accentset{\circ}{h}_m : \mathcal{X}^n \to \mathcal{H} \, ,
\end{equation*}
and defined as 
\begin{equation*}
\accentset{\circ}{h}_m (X_i) = \left( h_m \circ h_{m-1} \circ \cdots \circ h_2 \circ h_1 \right) (X_i) \, , \qquad i=1,2,\dots,n \, ,
\end{equation*} 
could be considered as a normalization of $\mathbf{X}_n$ where $\mathcal{H} \subset {\mathbb{R}^n}$. Throughout this work, we focus on a specific class of normalizations referred to as $m$-\textit{normalizations}, which contains every possible $(\mathscr{B}_{\mathcal{X}^n},\mathscr{B}_{\mathcal{H}})-$measurable function mapping the considered $\mathbf{X}_n$ into a strictly positive sample. 
\begin{definition}[$m$-normalization]
	The  $(\mathscr{B}_{\mathcal{X}^n},\mathscr{B}_{\mathcal{H}})-$measurable function $\accentset{\circ}{h}_m : \mathcal{X}^n \to \mathcal{H}$ is said to be an $m$-normalization of $\mathbf{X}_n$ if and only if its image satisfies $\mathcal{H} \subset {\mathbb{R}}_{>0}^{n}$.
\end{definition}

\begin{definition}[Discrete Lehmer Transform]
	\label{Thrm_Disc_Lehmer}
	The discrete Lehmer transform denoted by
	\begin{equation*}
	\breve{\mathscr{L}} : \overline{\mathbb{R}} \to \mathbb{R}_{>0} \, ,
	\end{equation*} is a
	$(\mathscr{B}_{\overline{\mathbb{R}}},\mathscr{B}_{\mathbb{R}_{>0}} )-$measurable function defined as
	\begin{align}
	\label{Eqn_Disc_Lehmer}
	\LT{n}{m}{\breve{s}} =
	\begin{cases} \max\limits_{i=1,\dots,n}\left\{\accentset{\circ}{h}_m(X_i)\right\} \, , & \quad	\mbox{if } \ \breve{s} = \infty \, , \\[1.0em]  
	\dfrac{\sum\limits_{i=1}^n \accentset{\circ}{h}^{\breve{s}}_m(X_i)}{\sum\limits_{i=1}^n \accentset{\circ}{h}^{\breve{s}-1}_m(X_i)} \, ,  &  \quad	\mbox{if } \ \breve{s} \in (-\infty,\infty) \, , \\[2em] 
	\min\limits_{i=1,\dots,n}\left\{\accentset{\circ}{h}_m(X_i)\right\} \, , & \quad	\mbox{if } \ \breve{s} = -\infty \, ,  \\
	\end{cases}
	\end{align}
	where $\accentset{\circ}{h}_m$ represents some $m$-normalization of the sample.
\end{definition}
According to Def. \ref{Thrm_Disc_Lehmer}, the discrete Lehmer transform $\LT{n}{m}{\breve{s}}$ maps every point $\breve{s} \in \overline{\mathbb{R}}$ into some statistic contained in the close interval 
\begin{equation*}
\left[ \min\limits_{i=1,\dots,n}\left\{\accentset{\circ}{h}_m(X_i)\right\} \, , \, \max\limits_{i=1,\dots,n}\left\{\accentset{\circ}{h}_m(X_i)\right\} \right] \, ,
\end{equation*}
where the points $\breve{s}$ (read $s$-breve) are referred to as breve moments of $\mathbf{X}_n$ under $m$-normalization $\accentset{\circ}{h}_m$. Table \ref{Table_Special_Cases} reports some of the widely-encountered breve moments in data analysis. Furthermore, for every statistic 
\begin{equation*}
T : \mathcal{X}^n \to \left[ \min\limits_{i=1,\dots,n}\left\{\accentset{\circ}{h}_m(X_i)\right\} \, , \, \max\limits_{i=1,\dots,n}\left\{\accentset{\circ}{h}_m(X_i)\right\} \right] \, ,
\end{equation*}
the following inverse image
\begin{equation*}
\breve{\mathscr{L}}^{-1} : \mathbb{R}_{>0} \to \overline{\mathbb{R}} \, ,
\end{equation*}
provides a set that contains the associated breve moment(s) of the sample under $m$-normalization $\accentset{\circ}{h}_m$ such that
\begin{equation*}
\LTinverse{n}{m}{T} = \left\{ \breve{s} \in \overline{\mathbb{R}} : \  
\LT{n}{m}{\breve{s}} = T \right\} \, .
\end{equation*}

\begin{table}[]
	\centering
	\caption{My caption}
	\label{Table_Special_Cases}
	\begin{tabular}{@{}cc@{}}
		\toprule
		\textbf{Breve moment domain} & \textbf{Sample domain}           \\ \midrule
		$-\infty$            & Minimum ($n\geq 1$)              \\
		$0$                  & Harmonic mean ($n\geq 1$)        \\
		$1/2$        & Geometric mean ($n = 2$)         \\
		$1$                  & Arithmetic mean ($n\geq 1$)      \\
		$2$                  & Contra-harmonic mean ($n\geq 1$) \\
		$\infty$            & Maximum ($n\geq 1$)              \\ \bottomrule
	\end{tabular}
\end{table}

\begin{lemma}
	For every injective $\accentset{\circ}{h}_m$,  the sufficient condition for $\breve{\mathscr{L}}$ to be
	\begin{enumerate}[(i)]
		\item a constant function is given by
		\begin{equation}
		\left\{ \exists \ i,j \in \left\{1,2.\cdots,n\right\} : i\neq j \implies \accentset{\circ}{h}_m(X_i) \neq \accentset{\circ}{h}_m(X_j) \right\} \stackrel{a.s.}{\to} 0 \, ;
		\end{equation}
		\item a monotone increasing function is given by
		\begin{equation}
		\label{Weak}
		\left\{ \exists \ i,j \in \left\{1,2.\cdots,n\right\} : i\neq j \implies \accentset{\circ}{h}_m(X_i) \neq \accentset{\circ}{h}_m(X_j) \right\} \stackrel{a.s.}{\to} 1 \, ;
		\end{equation}
		\item a strictly monotone increasing function is given by
		\begin{equation}
		\label{Strong}
		\left\{ \forall \ i,j \in \left\{1,2.\cdots,n\right\} : i\neq j \implies \accentset{\circ}{h}_m(X_i) \neq \accentset{\circ}{h}_m(X_j) \right\} \stackrel{a.s.}{\to} 1 \, .
		\end{equation}
	\end{enumerate}
\end{lemma}

\begin{lemma}
	For every injective $\accentset{\circ}{h}_m$,  the sufficient condition for $\breve{\mathscr{L}}$ to be
	\begin{enumerate}[(i)]
		\item a constant function is given by
		\begin{equation}
		\mathbb{P} \left[ \exists \ i,j \in \left\{1,2.\cdots,n\right\} : i\neq j \implies X_i \neq X_j \right] = 0 \, ;
		\end{equation}
		\item a monotone increasing function is given by
		\begin{equation}
		\mathbb{P} \left[ \exists \ i,j \in \left\{1,2.\cdots,n\right\} : i\neq j \implies X_i \neq X_j \right] = 1 \, ;
		\end{equation}
		\item a strictly monotone increasing function is given by
		\begin{equation}
		\mathbb{P} \left[ \forall \ i,j \in \left\{1,2.\cdots,n\right\} : i\neq j \implies X_i \neq X_j \right] = 1 \, .
		\end{equation}
	\end{enumerate}
\end{lemma}

\begin{lemma}
	The discrete Lehmer transform $\breve{\mathscr{L}}$ is differentiable at $\breve{s}_0$ in $P$-probability, and its $n$th order derivative is given by
	\begin{equation*}
	\dfrac{\partial^n}{\partial \breve{s}^n} \breve{\mathscr{L}} = 
	\breve{\mathscr{L}} \, \left[ \, \sum_{k=0}^{n} \dfrac{1}{k!} \left( \sum_{j=0}^{k} \left(-1\right)^j 
	{k \choose j} \,  \Lambda^{j} \, \dfrac{\partial^n}{\partial \breve{s}^n} \Lambda^{k-j} \right)   \right] \, ,
	\end{equation*}
	where
	\begin{equation*}
	\Lambda := \log{\breve{\mathscr{L}}} \, .
	\end{equation*}
\end{lemma}

\begin{theorem}[Inversion Theorem]
	\label{InverseLT}
	For a sample $\mathbf{X}_n$ having a (strictly) monotone increasing $\breve{\mathscr{L}}$, its discrete inverse Lehmer transform is a $(\mathscr{B}_{\mathbb{R}_{>0}}, \mathscr{B}_{\overline{\mathbb{R}}})-$measurable function defined as
	\begin{align}
	\breve{\mathscr{L}}^{-1}_{n}\{ \accentset{\circ}{h}_m \}(T) =
	\begin{cases} \infty \, , & 	\mbox{if } \ T = \max\limits_{i=1,\dots,n}\left\{\accentset{\circ}{h}_m(X_i)\right\} \, , \\[1.0em]  
	g(\breve{s}) \, ,  &  	\mbox{if } \ T\in \left( \min\limits_{i=1,\dots,n}\left\{\accentset{\circ}{h}_m(X_i)\right\} \, , \, \max\limits_{i=1,\dots,n}\left\{\accentset{\circ}{h}_m(X_i)\right\} \right) \, , \\[1.0em] 
	-\infty \, , & 	\mbox{if } \ T = \min\limits_{i=1,\dots,n}\left\{\accentset{\circ}{h}_m(X_i)\right\} \, , \\
	\end{cases}
	\end{align}
	where
	\begin{equation}
	g(\breve{s}) = \breve{s}_0 + \sum\limits_{k=1}^{\infty} \dfrac{\left(T-\breve{\mathscr{L}}_{n}\{ \accentset{\circ}{h}_m \}(\breve{s}_0)\right)^k }{k!} 
	\left\{ \lim_{\breve{s}\to \breve{s}_0} \left[ \dfrac{\partial^{k-1}}{\partial \breve{s}^{k-1}} \left( \dfrac{\breve{s}-\breve{s}_0}
	{\breve{\mathscr{L}}_{n}\{ \accentset{\circ}{h}_m \}(\breve{s})    -    \breve{\mathscr{L}}_{n}\{ \accentset{\circ}{h}_m \}(\breve{s}_0)}
	\right)^k \right] \right\} \, ,
	\end{equation}
	for some $\breve{s}_0\in \mathbb{R}$.
\end{theorem}


\begin{theorem}
	In a family $\mathcal{F}$ of linear transformations of some (strictly) monotone increasing $\breve{\mathscr{L}}$ where 
	\begin{equation}
	\label{Linear_Family}
	\mathcal{F} = \left\{ a + b \, \LTshort \, :  \, a\in \mathbb{R} \, , \, b\in \mathbb{R}_{>0} \right\} \, ,
	\end{equation}
	there exists one and only one $(\mathscr{B}_{\overline{\mathbb{R}}},\mathscr{B}_{\mathbb{R}})-$~measurable function $F(\breve{s})\in \mathcal{F}$ that satisfies the condition for being a distribution.
\end{theorem}

\begin{corollary}
	Let $\{A_k\}_{k\geq 1}$ and $\{B_k\}_{k\geq 1}$ be sequences of $(\mathscr{B}_{\mathcal{X}}, \mathscr{B}_{\mathbb{R}})-$
	and $(\mathscr{B}_{\mathcal{X}}, \mathscr{B}_{\mathbb{R}_{>0}})-$measurable functions such that $A_k \stackrel{P}{\to} A$ and $B_k \stackrel{P}{\to} B$, respectively. Then, for every (strictly) monotone increasing $\breve{\mathscr{L}}$, the following $(\mathscr{B}_{\overline{\mathbb{R}}},\mathscr{B}_{\mathbb{R}})-$measurable function
	\begin{equation*}
	F(\breve{s}) = A + B \, \LTshort(\breve{s}) \, ,
	\end{equation*}
	is a distribution if 
	\begin{equation*}
	A \stackrel{P}{\to} - \dfrac{\LTshort(-\infty)}{\LTshort(\infty)-\LTshort(-\infty)} \, ,
	\end{equation*}
	and
	\begin{equation*}
	B \stackrel{P}{\to} \dfrac{1}{\LTshort(\infty)-\LTshort(-\infty)} \, .
	\end{equation*}
\end{corollary}

\begin{corollary}
	For every (strictly) monotone increasing $\breve{\mathscr{L}}$, the following $(\mathscr{B}_{\overline{\mathbb{R}}},\mathscr{B}_{\mathbb{R}})-$measurable function
	\begin{equation*}
	F(\breve{s}) =  \LTshort(\breve{s}) \, ,
	\end{equation*}
	is a distribution if and only if 
	\begin{equation*}
	\LTshort(\infty) \stackrel{P}{\to} W_0(e) \, ,
	\end{equation*}
	and
	\begin{equation*}
	\LTshort(-\infty) \stackrel{P}{\to} W_0(0) \, .
	\end{equation*}
\end{corollary}

\begin{theorem}
	Assume the discrete Lehmer transform of a sample $\mathbf{X}_n$ satisfies the conditions $\LTshort(\infty) \stackrel{P}{\to} W_0(e)$ and $\LTshort(-\infty) \stackrel{P}{\to} W_0(0)$. Then, a probability density function for the extended real-valued random variable $\breve{S}$ could be derived using 
	
	\begin{equation}
	\label{pdf_linear}
	\begin{split}
	f_{\breve{S}} (\breve{s}; \alpha, \beta) \, = \, \, & \left(\dfrac{\partial}{\partial \breve{s}} \LTshort(\breve{s})\right) \,  \mathbb{I}_{(-\infty, \infty)}(\breve{s}) \\
	= \, \, 
	& \begin{multlined}[t] \left[ \sum_{i=1}^{n}  \sum_{k=i+1}^{n} 
	\left( \accentset{\circ}{h}_m(X_i) - \accentset{\circ}{h}_m(X_k) \right)
	\left( \log\accentset{\circ}{h}_m(X_i) - \log\accentset{\circ}{h}_m(X_k) \right)
	\left( \accentset{\circ}{h}_m(X_i) \, \accentset{\circ}{h}_m(X_k) \right)^{\breve{s}-1} \right] \\
	\times  { \left( \sum_{i=1}^{n} \accentset{\circ}{h}_m^{\breve{s}-1} (X_i) \right)^{-2}  } \, \mathbb{I}_{(-\infty, \infty)}(\breve{s}) \, , \end{multlined}
	\end{split}
	\end{equation}
	for every $\alpha\in \left(0,1\right]$ and $\beta \in \mathbb{R}_{>0}$.
\end{theorem}

\newpage
\begin{theorem}
	In a family $\mathcal{F}$ of non-linear transformations of some (strictly) monotone increasing $\breve{\mathscr{L}}$ where 
	\begin{equation}
	\label{Nonlinear_Family}
	\mathcal{F} = \left\{ a + b \, \LTshort^{1/\alpha} \, e^{\beta \LTshort} \, :  \, a\in \mathbb{R} \, , \, b\in \mathbb{R}_{>0} \, , \, \alpha\in \left(0,1\right] \, , \, \beta \in \mathbb{R}_{>0} \right\} \, ,
	\end{equation}
	there exists one and only one $(\mathscr{B}_{\overline{\mathbb{R}}},\mathscr{B}_{\mathbb{R}})-$measurable function $F(\breve{s})\in \mathcal{F}$ that satisfies the condition for being a distribution.
\end{theorem}

\begin{corollary}
	Let $\{A_k\}_{k\geq 1}$, $\{B_k\}_{k\geq 1}$ be sequences of $(\mathscr{B}_{\mathcal{X}}, \mathscr{B}_{\mathbb{R}})-$
	and $(\mathscr{B}_{\mathcal{X}}, \mathscr{B}_{\mathbb{R}_{>0}})-$measurable functions such that $A_k \stackrel{P}{\to} A$ and $B_k \stackrel{P}{\to} B$, respectively. Then, for every (strictly) monotone increasing $\breve{\mathscr{L}}$, the following $(\mathscr{B}_{\overline{\mathbb{R}}},\mathscr{B}_{\mathbb{R}})-$measurable function
	\begin{equation*}
	F(\breve{s}) = A + B \, \left(\LTshort(\breve{s})\right)^{1/\alpha} \,  e^{\beta \LTshort(\breve{s})} \, ,
	\end{equation*}
	is a distribution if 
	\begin{equation*}
	A \stackrel{P}{\to} - \, \dfrac{\exp\left\{\frac{1}{\alpha} \log\LTshort(-\infty) \, + \,  \beta \LTshort(-\infty)   \right\}}{ \exp\left\{\frac{1}{\alpha} \log\LTshort(\infty) \, + \,  \beta \LTshort(\infty)   \right\} - \,  \exp\left\{\frac{1}{\alpha} \log\LTshort(-\infty) \, + \,  \beta \LTshort(-\infty)\right\} }  \, ,
	\end{equation*}
	and
	\begin{equation*}
	B \stackrel{P}{\to} \dfrac{1}{ \exp\left\{\frac{1}{\alpha} \log\LTshort(\infty) \, + \,  \beta \LTshort(\infty)   \right\} \, - \,  \exp\left\{\frac{1}{\alpha} \log\LTshort(-\infty) \, + \,  \beta \LTshort(-\infty)   \right\} } \, ,
	\end{equation*}
	where $\alpha\in \left(0,1\right]$ and $\beta \in \mathbb{R}_{>0}$ denote some given parameters.
\end{corollary}

\begin{corollary}
	For every (strictly) monotone increasing $\breve{\mathscr{L}}$, the following $(\mathscr{B}_{\overline{\mathbb{R}}},\mathscr{B}_{\mathbb{R}})-$measurable function
	\begin{equation*}
	F(\breve{s}) =  \left(\LTshort(\breve{s})\right)^{1/\alpha} \,  e^{\beta \LTshort(\breve{s})} \, ,
	\end{equation*}
	is a distribution if and only if 
	\begin{equation*}
	\LTshort(\infty) \stackrel{P}{\to} \dfrac{W_0(\alpha \beta)}{\alpha \beta} \, ,
	\end{equation*}
	and
	\begin{equation*}
	\LTshort(-\infty) \stackrel{P}{\to} W_0(0) \, ,
	\end{equation*}
	where $\alpha\in \left(0,1\right]$ and $\beta \in \mathbb{R}_{>0}$ denote some given parameters.
\end{corollary}

\begin{theorem}[Breve Distribution]
	Assume the discrete Lehmer transform of a sample $\mathbf{X}_n$ satisfies the conditions $\LTshort(\infty) \stackrel{P}{\to} \frac{W_0(\alpha \beta)}{\alpha \beta}$ and $\LTshort(-\infty) \stackrel{P}{\to} W_0(0)$. Then, an extended real-valued random variable $\breve{S}$ is said to follow a Breve distribution if its probability density function is given by
	\begin{equation}
	\label{pdf_nonlinear}
	\begin{split}
	f_{\breve{S}} (\breve{s}; \alpha, \beta) \, = \, \, &  \left(\dfrac{\partial}{\partial \breve{s}} \LTshort(\breve{s})\right) \, \dfrac{1}{\alpha} \, \exp{\{\beta \LTshort(\breve{s}) \}} \,
	\left( 1+ \alpha \, \beta \,  \LTshort(\breve{s}) \right) \, \left( \LTshort(\breve{s}) \right)^{\frac{1}{\alpha}-1} \, \mathbb{I}_{(-\infty, \infty)}(\breve{s})  \\
	= & \begin{multlined}[t]\left[ \sum_{i=1}^{n}  \sum_{k=i+1}^{n} 
	\left( \accentset{\circ}{h}_m(X_i) - \accentset{\circ}{h}_m(X_k) \right)
	\left( \log\accentset{\circ}{h}_m(X_i) - \log\accentset{\circ}{h}_m(X_k) \right)
	\left( \accentset{\circ}{h}_m(X_i) \, \accentset{\circ}{h}_m(X_k) \right)^{\breve{s}-1} \right] \\
	\qquad \qquad \quad \times  { \left( \sum_{i=1}^{n} \accentset{\circ}{h}_m^{\breve{s}-1} (X_i) \right)^{-2}  } \,
	\dfrac{1}{\alpha} \, \exp\left\{ {\frac{\beta \sum\limits_{i=1}^n \accentset{\circ}{h}^{\breve{s}}_m(X_i)}{\sum\limits_{i=1}^n \accentset{\circ}{h}^{\breve{s}-1}_m(X_i)}} \right\} \,
	\left( 1 + \alpha \, \beta \, \dfrac{\sum\limits_{i=1}^n \accentset{\circ}{h}^{\breve{s}}_m(X_i)}{\sum\limits_{i=1}^n \accentset{\circ}{h}^{\breve{s}-1}_m(X_i)} \right)  \\
	\quad \times  \, \left( \dfrac{\sum\limits_{i=1}^n \accentset{\circ}{h}^{\breve{s}}_m(X_i)}{\sum\limits_{i=1}^n \accentset{\circ}{h}^{\breve{s}-1}_m(X_i)} \right)^{\frac{1}{\alpha}-1}
	\mathbb{I}_{(-\infty, \infty)}(\breve{s}) \, , \end{multlined}
	\end{split}
	\end{equation}
	for every $\alpha\in \left(0,1\right]$ and $\beta \in \mathbb{R}_{>0}$. A Breve random variable having the probability density function () is denoted by $\breve{S} \sim Breve(\breve{s}; \alpha, \beta)$.
\end{theorem}

\begin{theorem}
	The extreme points of (\ref{pdf_nonlinear}) satisfy the following relation
	\begin{equation}
	\LTshort(\breve{s}) = \dfrac{1}{\alpha \beta} \, W_0\left( \, \alpha \beta \left(\dfrac{C_1}{\alpha} \left(\breve{s}+ C_2\right)\right)^\alpha \,  \right) \, ,
	\end{equation}
	for some constants $C_1$ and $C_2$ such that
	\begin{equation*}
	\LTshort(\infty) \stackrel{P}{\to} \dfrac{W_0(\alpha \beta)}{\alpha \beta} \, ,
	\end{equation*}
	and
	\begin{equation*}
	\LTshort(-\infty) \stackrel{P}{\to} W_0(0) \, .
	\end{equation*}
\end{theorem}

\begin{theorem}[Log-Breve Distribution]
	Assume the discrete Lehmer transform of a sample $\mathbf{X}_n$ satisfies the conditions $\LTshort(\infty) \stackrel{P}{\to} \exp\{\frac{W_0(\alpha \beta)}{\alpha \beta}\}$ and $\LTshort(-\infty) \stackrel{P}{\to} \exp\{W_0(0)\}$. Then, an extended real-valued random variable $\breve{S}$ is said to follow a Log-Breve distribution if its probability density function is given by
	\begin{equation}
	\label{pdf_nonlinear}
	\begin{split}
	f_{\breve{S}} (\breve{s}; \alpha, \beta) \, = \, \, & \begin{multlined}[t]\left[ \sum_{i=1}^{n}  \sum_{k=i+1}^{n} 
	\left( \accentset{\circ}{h}_m(X_i) - \accentset{\circ}{h}_m(X_k) \right)
	\left( \log\accentset{\circ}{h}_m(X_i) - \log\accentset{\circ}{h}_m(X_k) \right)
	\left( \accentset{\circ}{h}_m(X_i) \, \accentset{\circ}{h}_m(X_k) \right)^{\breve{s}-1} \right] \\
	\times  { \left( \sum_{i=1}^{n} \accentset{\circ}{h}_m^{\breve{s}-1} (X_i) \right)^{-2}  } \,
	\dfrac{1}{\alpha} \, \exp\left\{ {\frac{\beta \sum\limits_{i=1}^n \accentset{\circ}{h}^{\breve{s}}_m(X_i)}{\sum\limits_{i=1}^n \accentset{\circ}{h}^{\breve{s}-1}_m(X_i)}} \right\} \,
	\left( 1 + \alpha \, \beta \, \dfrac{\sum\limits_{i=1}^n \accentset{\circ}{h}^{\breve{s}}_m(X_i)}{\sum\limits_{i=1}^n \accentset{\circ}{h}^{\breve{s}-1}_m(X_i)} \right) 
	\left( \dfrac{\sum\limits_{i=1}^n \accentset{\circ}{h}^{\breve{s}}_m(X_i)}{\sum\limits_{i=1}^n \accentset{\circ}{h}^{\breve{s}-1}_m(X_i)} \right)^{\frac{1}{\alpha}-1}  \\
	\times  \, 
	\left( \exp\left\{\frac{W_0(\alpha \beta)}{\alpha^2 \beta} \, + \,  \beta \exp\{\frac{W_0(\alpha \beta)}{\alpha \beta}\}   \right\} \, - \,  \exp\left\{ \beta    \right\} \right)^{-1}
	\mathbb{I}_{(-\infty, \infty)}(\breve{s}) \, , \end{multlined}
	\end{split}
	\end{equation}
	for every $\alpha\in \left(0,1\right]$ and $\beta \in \mathbb{R}_{>0}$. A Log-Breve random variable having the probability density function () is denoted by $\breve{S} \sim LogBreve(\breve{s}; \alpha, \beta)$.
\end{theorem}


\end{document}